\documentclass{article}
\usepackage{arxiv}
\usepackage[utf8]{inputenc} 
\usepackage[T1]{fontenc}    
\usepackage{hyperref}       
\usepackage{url}            
\usepackage{booktabs}       
\usepackage{amsfonts}       
\usepackage{nicefrac}       
\usepackage{microtype}      
\usepackage{amsmath}
\usepackage{amssymb}
\usepackage{mathtools}
\usepackage{graphicx}
\usepackage{listings}
\usepackage{appendix}
\usepackage[square,sort,comma,numbers]{natbib}
\graphicspath{ {./images/} }
\DeclareMathOperator{\atantwo}{atan2}
\title{Recurrent Attention Model with Log-Polar Mapping is Robust against Adversarial Attacks}

\author{
  Taro Kiritani\\
  ExaWizards Inc.\\
  \texttt{taro.kiritani@exwzd.com} \\
   \And
 Koji Ono\\
  ExaWizards Inc.\\
  \texttt{koji.ono@exwzd.com} \\
}

\begin{document}
\maketitle

\begin{abstract}
Convolutional neural networks are vulnerable to small $\ell^p$ adversarial attacks, while the human visual system is not. Inspired by neural networks in the eye and the brain, we developed a novel artificial neural network model that recurrently collects data with a log-polar field of view that is controlled by attention. We demonstrate the effectiveness of this design as a defense against SPSA and PGD adversarial attacks. It also has beneficial properties observed in the animal visual system, such as reflex-like pathways for low-latency inference, fixed amount of computation independent of image size, and rotation and scale invariance. The code for experiments is available at \url{https://gitlab.com/exwzd-public/kiritani_ono_2020}.
\end{abstract}

\section{Introduction}
Convolutional neural networks (CNNs) were designed after the hierarchical animal visual system \cite{fukushima1980neocognitron} discovered by Hubel and Wiesel \cite{hubel1962receptive}. However, CNNs represent visual inputs very differently from the animal brain. In standard, feedforward CNNs trained on ImageNet \cite{imagenet_cvpr09}, a unit in a convolution layer has a small effective receptive field \cite{luo2016understanding}. CNNs, as a result, have difficulty in capturing shapes of large objects, and are biased toward local texture \cite{geirhos2018imagenet, brendel2018approximating}. The existence of adversarial examples \cite{szegedy2013intriguing} further highlights CNNs' weakness and difference from the human visual system. Small $\ell^p$ perturbation of images can fool the human brain when images are presented to subjects briefly for 63-71 ms; however, visual recognition of humans is not disrupted if subjects can see perturbed images long enough \cite{elsayed2018adversarial}. In this study, we aim to emulate the robustness of natural visual systems.

The animal visual system is a recurrent neural network (RNN) that actively collects data over time \cite{hofer2011differential}. On the human retina, the acuity is highest in the fovea with densely packed photoreceptors, and the resolution decreases as a function of eccentricity \cite{curcio1990human}. Rapid eye movements, and the non-uniform resolution in the retina allow animals to efficiently scrutinize informative regions of the visual scene while processing other areas at lower resolution \cite{land1990eye, marshall2014shrimps}. In the mammalian neocortex, the signal from the eye is routed to two parallel data streams \cite{mishkin1983object, goodale1992separate, wang2011gateways}. In the ventral, or "what" pathway, the identity of visual objects is extracted. The dorsal, or "where" pathway on the other hand, is involved with spatial awareness and localization of objects. We take inspiration from this sampling strategy of the brain and propose a novel computer vision model, Recurrent Attention Model with Log Polar Mapping (RAM-LPM). RAM-LPM has a log-polar field of view (FOV) that is mapped to a regular grid-like tensor. The tensor is then processed in "what" and "where" RNNs. The "what" pathway is trained to learn the features of patches for object classification, whereas the movement of FOV is controlled by the "where" pathway.

We show that RAM-LPM is resistant to simultaneous perturbation stochastic approximation (SPSA) \cite{uesato2018adversarial} and projected gradient descent (PGD) \cite{madry2017towards} attacks. We also discuss other desirable properties of RAM-LPM. Similar to the animal eye, the log-polar FOV is highly invariant to scaling and rotation. RAM-LPM also responds to inputs quickly with a reflex circuit. Another advantage is the fixed amount of computation that is independent of image size; in standard CNNs, the amount of computation linearly scales to pixel number. RAM-LPM thus overcomes some of the shortcomings of CNNs, demonstrating the effectiveness of bridging neuroscience and deep learning research.

\section{Background and Related Works}
\subsection{Log-Polar Vision}
The eye does not process the entire visual field evenly. Photoreceptive cells in the human eye are abundant around the central region called the fovea, whereas the density decreases in the periphery \cite{curcio1990human, finlay2008number, inzunza1991topography}. The development of FOV in a log-polar coordinate system was motivated by the distribution of photoreceptors in the eye \cite{massone1985form, messner1985image, traver2010review}. The advantages of this sampling method, compared to sampling with a conventional Cartesian lattice, include efficient compression of information, as well as rotation and scale equivalence \cite{wilson1992pattern, esteves2017polar}. The brain's ability to perceive objects is also largely unaffected by rotation and scaling \cite{guyonneau2006animals, isik2013dynamics}. Anatomical and physiological studies demonstrated that the projection from the retina to the primary visual cortex can be approximated by log-polar mapping \cite{daniel1961representation, schwartz1977spatial, tootell1982deoxyglucose}, suggesting that the log-polar FOV captures the computation in the mammalian visual system better than the Cartesian FOV \cite{zetzsche1999atoms}.

\subsection{Attention}
The movement of the eye, together with the non-uniform spatial resolution in the eye, further enables efficient sampling in the visual field \cite{kayser2006fixations, elazary2008interesting, Schutz7547}. In computer vision, hard attention, a mechanism that selects and processes only small portions of an image was inspired by the animal visual system. Hard attention is used to boost the performance as well as to reduce the amount of computation \cite{mnih2014recurrent, elsayed2019saccader}.

\subsection{Reflex Circuit}
Animals respond to sensory inputs over variable time scale. The fastest output is mediated by reflex in the peripheral nervous system \cite{sherrington1910flexion}. Slower but more thoughtful responses involve recurrent circuits in the central nervous system \cite{kubota1971prefrontal,inagaki2019discrete}. Recent neural networks employ short-cut connections; however, they are intended for effective backpropagation of derivatives of very deep neural networks \cite{he2016deep}, and not used for responses with short latency.

\subsection{Adversarial Attacks}
An adversarial attack is a technique to find a perturbation that changes the prediction of a machine learning model \cite{szegedy2013intriguing}. The perturbation can be very small and imperceptible to human eyes. In this study, we only consider perturbation of small $\ell^p$ norm added to original images. The human visual system is perturbed much less than CNNs if subjects are allowed to see images long enough \cite{elsayed2018adversarial}. Luo et al. \cite{luo2015foveation} suggested foveation as a defense to adversarial attacks, but Athaye et al \cite{pmlr-v80-athalye18b} generated examples that remain adversarial after scaling, translation and other changes in viewpoint. Furthermore, hard attention with square crops was recently shown to be an inadequate defence \cite{elsayed2019saccader}. Rotation-equivalent networks were shown to be robust to spatially transformed adversarial attacks \cite{xiao2018spatially}; however, they were still vulnerable to $\ell^p$ adversarial attacks \cite{dumont2018robustness}.


\section{Model Formulation}
\begin{figure}
\includegraphics[width=\textwidth]{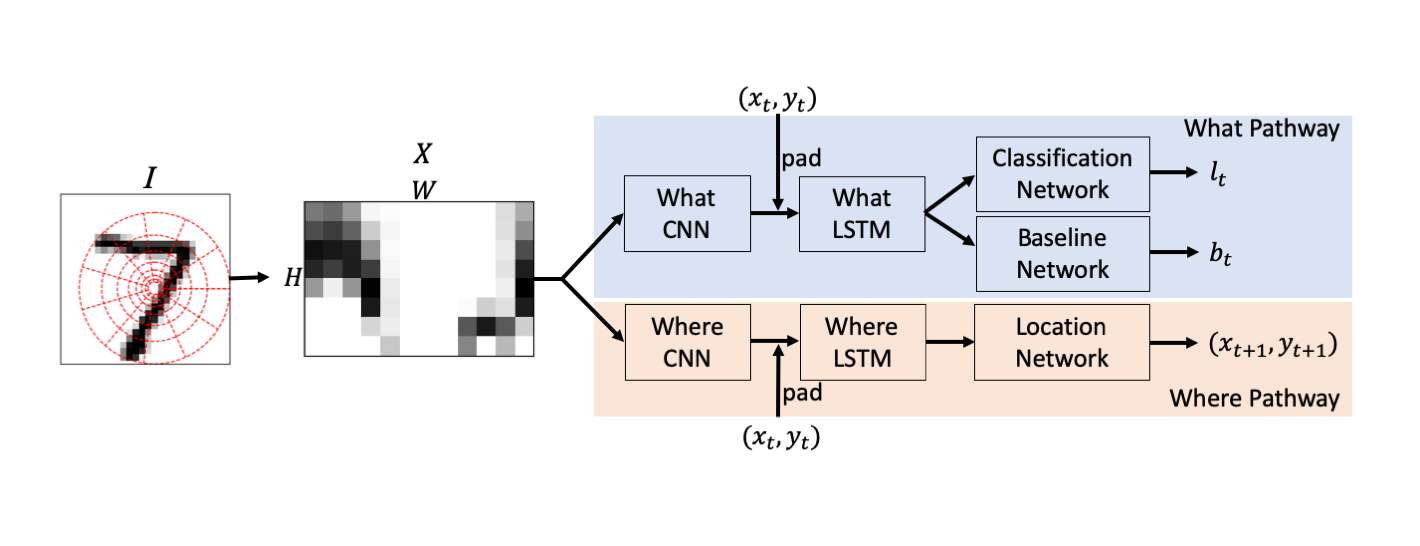}
\caption{\textbf{Architecutre of RAM-LPM}. A patch bounded by two concentric circles around $(x_{t}, y_{t})$ in image $I$ is mapped to a tensor with a shape of $[H,W,C]$. This tensor is processed in two separate pathways; the identity of the image $I$ is learned in "what" pathway, and "where" pathway computes the position for the foveation in the next time step.}
\label{fig:fig1}
\end{figure}

Like the recurrent attention model by \cite{mnih2014recurrent}, RAM-LPM has access to a fraction of an image at each time step. It learns both the identity of images, and the policy on where to attend in an image based on previous patches, and actions. The model consists of subnetworks that are summarized in figure \ref{fig:fig1}. At each time step $t$, a small patch around $(x_{t}, y_{t})$ is received by the model, and this patch is routed to "what" and "where" pathways. The logits $l_{t}$, location of the next foveation $(x_{t+1}, y_{t+1})$, and the prediction of reward $b_t$ (discussed later) are computed. The implementations of the sub-networks are discussed below.
\subsection{Sampling Method}
A patch of an image is extracted in a similar way described in \cite{esteves2017polar}. We consider an image $I$ in a 2D Cartesian coordinate system. Let $f(x,y,c)$ be the interpolated pixel intensity at $(x,y)$ in $I$ for channel $c$, and $(x_{t}, y_{t})$ be a point in $I$ that is learned by the location network. A point $(x,y)$ in $I$ is mapped to a point $(\rho, \theta)$ in a log-polar coordinate system such that,

\begin{equation}
\rho = \ln{\sqrt{(x-x_{t})^{2} + (y-y_{t})^{2}}},
\end{equation}
and 
\begin{equation}
\theta = \atantwo(y-y_{0}, x-x_{0}).
\end{equation}

Let $g(\rho, \theta, c)$ be the pixel intensity at $(\rho, \theta)$ for channel $c$ in log-polar coordinates:

\begin{equation}
    g(\rho, \theta, c) = f(e^{\rho} \cos{\theta}, e^{\rho} \sin{\theta}, c) = f(x,y,c).
\end{equation}

A region bounded by two concentric circles in $I$ is the field of view. In log-polar coordinates, the FOV is defined by
\begin{equation}
    \rho_{min} < \rho < \rho_{max},
\end{equation}
and
\begin{equation}
    -\pi \leq \theta < \pi.
\end{equation}
The FOV is evenly divided in log-polar coordinates into $H$ by $W$ sections. A tensor $X$ with shape $[H, W, C]$ whose elements are given by

\begin{equation}
    X[h,w,c]=g(\frac{(h-1)\rho_{max}-(H-h)\rho_{min}}{H-1},-\pi + \frac{2\pi w}{W},c)
\end{equation}

is fed to "what" and "where" pathways.

\subsection{What Pathway}
$X$ is first processed by a feedforward CNN network. Since $X$ is contiguous about the angular axis in the original image $I$, we use wrap-around padding \cite{esteves2017polar} before convolution and max pooling layers. The output of the CNN is flattened, padded with the ouput of the location network $(x_{t}, y_{t})$, and passed to the following LSTM network. The classification network and baseline network are both fully connected networks that output the logits for classification, and baseline used in REINFORCE training, respectively. We minimize the cross entropy loss by backpropagating the derivatives through the classification network, the LSTM network and CNN of the what pathway. The parameters in the baseline network are trained by minimizing the MSE loss from $b_{t}$ and reward $R$; $R=1$ when the prediction is correct; and $R=0$ otherwise. Adam optimizer \cite{kingma2014adam} is used to update the parameters.

\subsection{Where Pathway}
As in the "what" pathway, $X$ is first processed by a CNN. The flattened output padded with $(x_{t}, y_{t})$, is recurrently fed to a LSTM network. The location network consists of fully connected layers followed by parameterized Gaussian random number generators. The fully connected layers output a pair of real numbers $(\mu_{x}, \mu_{y})$ that are used to generate the position of the FOV at the next time step:
\begin{equation}
x_{t+1} \sim N(\mu_{x}, \sigma),
\end{equation}
and
\begin{equation}
y_{t+1} \sim N(\mu_{y}, \sigma).
\end{equation}
$\sigma$ could also be leaned is fixed in our experiments. The parameters in this pathway were updated using the REINFORCE algorithm with baseline subtraction \cite{mnih2014recurrent}.
The initial position of the FOV is randomized with uniform distributions:
\begin{equation}
    x_{0} \sim \mathcal{U}(-0.25, 0.25),
\end{equation}
and
\begin{equation}
    y_{0} \sim \mathcal{U}(-0.25, 0.25),
\end{equation}
where the center of the image is $(0, 0)$ and the length of the larger side of the image is 2.

\subsection{Training and Inference}
The model samples patches $glimpseNum$ times, and the final logit $l_{glimpseNum}$ is used in the cross entropy loss function in the "what" pathway. During inference, the forward pass is repeated $mcSample$ times, and the logits for each class are averaged.

\section{Experiments}
\subsection{Training on ImageNet} \label{Training on ImageNet}
\begin{figure}[h]
\includegraphics[width=\textwidth]{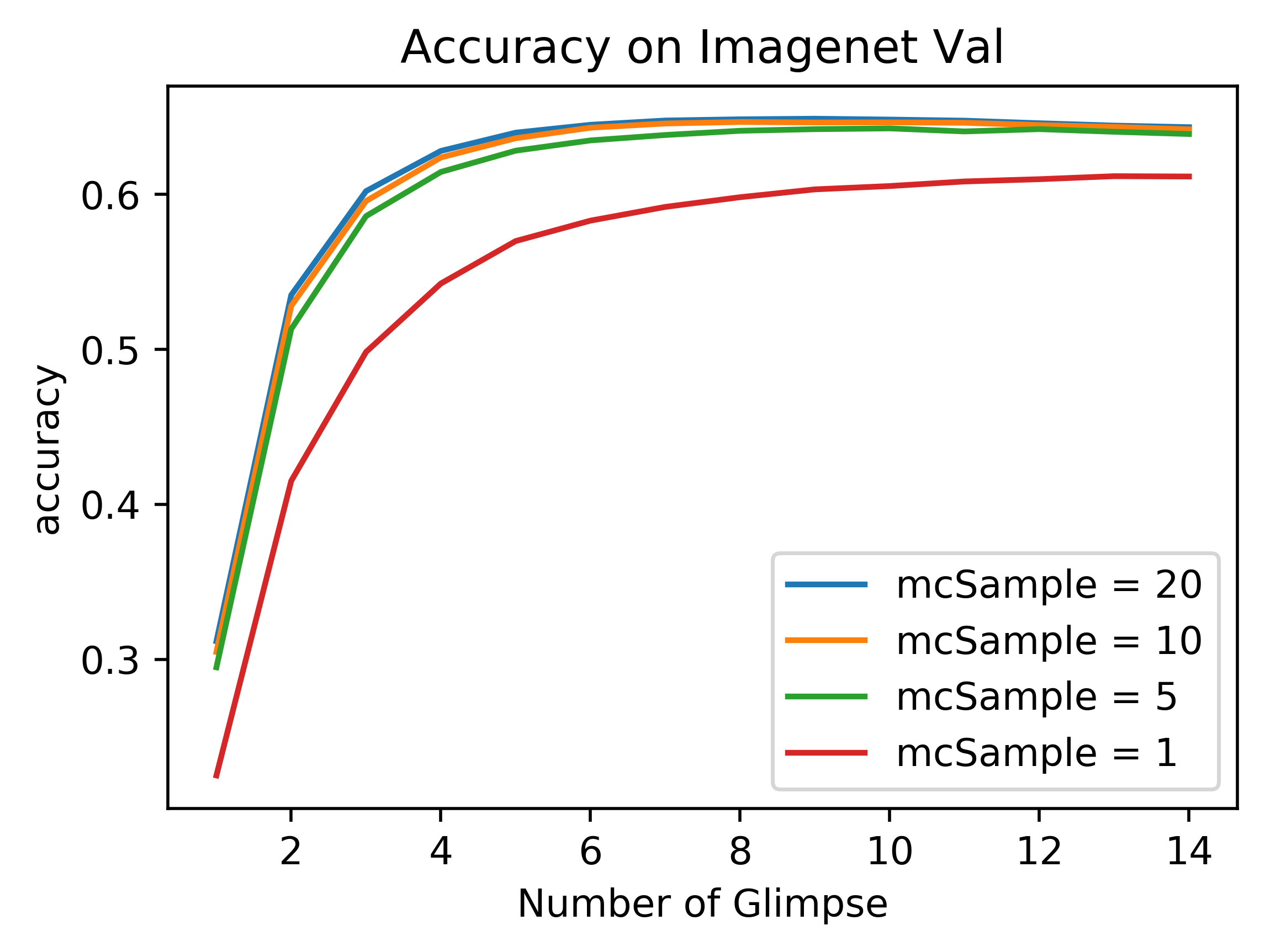}
\caption{\textbf{Prediction performance on Imagenet validation dataset.} Accuracy of the trained model as a function of the number of glimpses and the number of forward paths. Larger $glimpseNum$ and $mcSample$ led to higher accuracies. With $glimpseNum=10$ and $mcSample=20$, the accuracy was 64.8\%}
\label{fig:inference}
\end{figure}

We trained RAM-LPM on the Imagenet dataset \cite{ILSVRC15}. To avoid information loss caused by cropping and shrinking, the images are first zero-padded either vertically and horizontally to make them a square. The images are then resized to 512 x 512 pixels. Note that the amount of computation for RAM-LPM does not depend on the number of pixels in an image. The model performance on the validation set is summarized in figure \ref{fig:inference}. The fast reflex-like output is followed by slower response which is more accurate. Larger $mcSample$ also increases the accuracy.

\subsection{Adversarial Attacks}
The robustness of the trained model was tested on 300 or 1000 images randomly picked from ImageNet validation dataset, for SPSA and PGD attacks respectively. The gradient computed in "what" pathway of RAM-LPM with respect to the input was used in the PGD attack \cite{elsayed2019saccader}. We preprocessed the images in two different ways since the difficulty of finding an adversarial example can depend on image size \cite{uesato2018adversarial}. In the first preparation, we evaluated the robustness without cropping and resizing. Instead, zeros were padded horizontally or vertically to make the images a square. The padded regions were not changed by the adversarial attacks. In the second preparation, the short side of images was resized to 224 pixels preserving the aspect ratio, and then the central portion of 224 x 224 pixels was cropped. The images classified correctly by the trained model with $glimpseNum = 10$ and $mcSample = 1$ were perturbed by untargeted $\ell^{\infty}$ SPSA and PGD attacks implemented in advertorch \cite{ding2019advertorch}. For SPSA attack, we used $\epsilon=2/255$, perturbation size $\delta=0.01$, maximum iterations of $100$, batch size of $ 8192$, and Adam learning rate of $0.01$. These parameters are the same as in \cite{uesato2018adversarial}. For PGD attack, we used $\epsilon=2/255$, step size of $0.2/255$, and maximum iterations of $300$. We used the parameters specified in Appendix D of \cite{elsayed2019saccader} except for the step size which was decreased from $0.5/255$. This increased the success rate of the PGD attack.

The number of failures and successes of the attacks are summarized in Table \ref{tab:adversarial}. RAM-LPM was resistant to both attacks especially when the image sizes were not changed.

\begin{table}[h]
\caption{\textbf{Results of Adversarial Attack Experiments.} Only correctly classified images are further perturbed by the attacks.}
\centering
\begin{tabular}{l c c c c}
&SPSA (224 x 224)&SPSA (original size)&PGD (224 x 224)&PGD (original size)\\ 
\hline
Success : Failure : Incorrect &55 : 101 : 144&17 : 159 : 124&436 : 62 : 502& 269 : 290 : 441\\
\hline
\end{tabular}
\label{tab:adversarial}
\end{table}

\subsection{Rotation, Scale and Translation Invariance} \label{Rotation and Scale Invariance}

We evaluated the performance RAM-LPM on SIM2MNIST \cite{esteves2017polar} dataset. SIM2MNIST is a MNIST variant that is perturbed with clutters, rotation, scaling and translation. Table \ref{tab:mnist_var} shows our result along with the performance of other models reported in \cite{esteves2017polar}. The performance of RAM-LPM is comparable to that of PTNs, suggesting RAM-LPM achieves representation invariance to various transformations.

\begin{table}
\caption{\textbf{Error rate on SIM2MNIST test dataset.}}
\centering
\begin{tabular}{l c}
Model & Error (\%)\\ 
\hline
HNet \cite{worrall2017harmonic} implementation in \cite{esteves2017polar}& 9.28\\
PTN-S \cite{esteves2017polar} & 5.44\\
PTN-B \cite{esteves2017polar} & 5.03\\
RAM-LPM (ours)& 5.00\\
STN-B \cite{jain2015comparative} implementation in \cite{esteves2017polar}& 12.35\\
\hline
\end{tabular}

\label{tab:mnist_var}
\end{table}

\section{Conclusion}
In this work, we propose a novel approach for image classification that combines log-polar mapping and hard attention. We demonstrated that RAM-LPM achieves robustness against SPSA and PGD attacks. Robustness of brain-like models corroborates the notion that adversarial examples are human-centric phenomena, and suggests that RAM-LPM learns "robust features" \cite{ilyas2019adversarial}. Other beneficial properties of RAM-LPM include amount of computation independent of image size, reflex-like fast response, and invariance to rotation, scaling and translation. We hope our approach demonstrated the effectiveness of applying signal processing mechanisms in the brain to deep learning models. We expect our approach to be useful in safety-critical applications or analysis of high-resolution microscopy/satellite images where rotation/scale/translation invariance is exhibited.

\section*{Acknowledgement}
We are grateful to Jonathan Hough of ExaWizards Inc. for helpful feedback on the manuscript.

\bibliographystyle{alpha}
\bibliography{ms}

\newcommand{\etalchar}[1]{$^{#1}$}
\begin{thebibliography}{UOvdOK18}

\bibitem[AEIK18]{pmlr-v80-athalye18b}
Anish Athalye, Logan Engstrom, Andrew Ilyas, and Kevin Kwok.
\newblock Synthesizing robust adversarial examples.
\newblock In Jennifer Dy and Andreas Krause, editors, {\em Proceedings of the
  35th International Conference on Machine Learning}, volume~80 of {\em
  Proceedings of Machine Learning Research}, pages 284--293, Stockholmsmässan,
  Stockholm Sweden, 10--15 Jul 2018. PMLR.

\bibitem[BB19]{brendel2018approximating}
Wieland Brendel and Matthias Bethge.
\newblock Approximating {CNN}s with bag-of-local-features models works
  surprisingly well on imagenet.
\newblock In {\em International Conference on Learning Representations}, 2019.

\bibitem[CSKH90]{curcio1990human}
Christine~A Curcio, Kenneth~R Sloan, Robert~E Kalina, and Anita~E Hendrickson.
\newblock Human photoreceptor topography.
\newblock {\em Journal of comparative neurology}, 292(4):497--523, 1990.

\bibitem[DDS{\etalchar{+}}09]{imagenet_cvpr09}
J.~Deng, W.~Dong, R.~Socher, L.-J. Li, K.~Li, and L.~Fei-Fei.
\newblock {ImageNet: A Large-Scale Hierarchical Image Database}.
\newblock In {\em CVPR09}, 2009.

\bibitem[DMM18]{dumont2018robustness}
Beranger Dumont, Simona Maggio, and Pablo Montalvo.
\newblock Robustness of rotation-equivariant networks to adversarial
  perturbations.
\newblock {\em arXiv preprint arXiv:1802.06627}, 2018.

\bibitem[DW61]{daniel1961representation}
PM~Daniel and D~Whitteridge.
\newblock The representation of the visual field on the cerebral cortex in
  monkeys.
\newblock {\em The Journal of physiology}, 159(2):203--221, 1961.

\bibitem[DWJ19]{ding2019advertorch}
Gavin~Weiguang Ding, Luyu Wang, and Xiaomeng Jin.
\newblock {AdverTorch} v0.1: An adversarial robustness toolbox based on
  pytorch.
\newblock {\em arXiv preprint arXiv:1902.07623}, 2019.

\bibitem[EABZD17]{esteves2017polar}
Carlos Esteves, Christine Allen-Blanchette, Xiaowei Zhou, and Kostas
  Daniilidis.
\newblock Polar transformer networks.
\newblock {\em arXiv preprint arXiv:1709.01889}, 2017.

\bibitem[EI08]{elazary2008interesting}
Lior Elazary and Laurent Itti.
\newblock Interesting objects are visually salient.
\newblock {\em Journal of vision}, 8(3):3--3, 2008.

\bibitem[EKL19]{elsayed2019saccader}
Gamaleldin Elsayed, Simon Kornblith, and Quoc~V Le.
\newblock Saccader: Improving accuracy of hard attention models for vision.
\newblock In {\em Advances in Neural Information Processing Systems}, pages
  700--712, 2019.

\bibitem[ESC{\etalchar{+}}18]{elsayed2018adversarial}
Gamaleldin~F. Elsayed, Shreya Shankar, Brian Cheung, Nicolas Papernot, Alex
  Kurakin, Ian Goodfellow, and Jascha Sohl-Dickstein.
\newblock Adversarial examples that fool both computer vision and time-limited
  humans, 2018.

\bibitem[FFY{\etalchar{+}}08]{finlay2008number}
Barbara~L Finlay, Edna Cristina~S Franco, Elizabeth~S Yamada, Justin~C Crowley,
  Michael Parsons, Jos{\'e} Augusto~PC Muniz, and Luiz Carlos~L Silveira.
\newblock Number and topography of cones, rods and optic nerve axons in new and
  old world primates.
\newblock {\em Visual Neuroscience}, 25(3):289--299, 2008.

\bibitem[Fuk80]{fukushima1980neocognitron}
Kunihiko Fukushima.
\newblock Neocognitron: A self-organizing neural network model for a mechanism
  of pattern recognition unaffected by shift in position.
\newblock {\em Biological cybernetics}, 36(4):193--202, 1980.

\bibitem[GKT06]{guyonneau2006animals}
Rudy Guyonneau, Holle Kirchner, and Simon~J Thorpe.
\newblock Animals roll around the clock: The rotation invariance of ultrarapid
  visual processing.
\newblock {\em Journal of Vision}, 6(10):1--1, 2006.

\bibitem[GM92]{goodale1992separate}
Melvyn~A Goodale and A~David Milner.
\newblock Separate visual pathways for perception and action.
\newblock {\em Trends in neurosciences}, 15(1):20--25, 1992.

\bibitem[GRM{\etalchar{+}}18]{geirhos2018imagenet}
Robert Geirhos, Patricia Rubisch, Claudio Michaelis, Matthias Bethge, Felix~A
  Wichmann, and Wieland Brendel.
\newblock Imagenet-trained cnns are biased towards texture; increasing shape
  bias improves accuracy and robustness.
\newblock {\em arXiv preprint arXiv:1811.12231}, 2018.

\bibitem[HKP{\etalchar{+}}11]{hofer2011differential}
Sonja~B Hofer, Ho~Ko, Bruno Pichler, Joshua Vogelstein, Hana Ros, Hongkui Zeng,
  Ed~Lein, Nicholas~A Lesica, and Thomas~D Mrsic-Flogel.
\newblock Differential connectivity and response dynamics of excitatory and
  inhibitory neurons in visual cortex.
\newblock {\em Nature neuroscience}, 14(8):1045, 2011.

\bibitem[HW62]{hubel1962receptive}
David~H Hubel and Torsten~N Wiesel.
\newblock Receptive fields, binocular interaction and functional architecture
  in the cat's visual cortex.
\newblock {\em The Journal of physiology}, 160(1):106--154, 1962.

\bibitem[HZRS16]{he2016deep}
Kaiming He, Xiangyu Zhang, Shaoqing Ren, and Jian Sun.
\newblock Deep residual learning for image recognition.
\newblock In {\em Proceedings of the IEEE conference on computer vision and
  pattern recognition}, pages 770--778, 2016.

\bibitem[IBSA91]{inzunza1991topography}
Oscar Inzunza, Hermes Bravo, Ricardo~L Smith, and Manuel Angel.
\newblock Topography and morphology of retinal ganglion cells in falconiforms:
  A study on predatory and carrion-eating birds.
\newblock {\em The Anatomical Record}, 229(2):271--277, 1991.

\bibitem[IFRS19]{inagaki2019discrete}
Hidehiko~K Inagaki, Lorenzo Fontolan, Sandro Romani, and Karel Svoboda.
\newblock Discrete attractor dynamics underlies persistent activity in the
  frontal cortex.
\newblock {\em Nature}, 566(7743):212--217, 2019.

\bibitem[IMLP13]{isik2013dynamics}
Leyla Isik, Ethan~M Meyers, Joel~Z Leibo, and Tomaso~A Poggio.
\newblock The dynamics of invariant object recognition in the human visual
  system.
\newblock {\em American Journal of Physiology-Heart and Circulatory
  Physiology}, 2013.

\bibitem[IST{\etalchar{+}}19]{ilyas2019adversarial}
Andrew Ilyas, Shibani Santurkar, Dimitris Tsipras, Logan Engstrom, Brandon
  Tran, and Aleksander Madry.
\newblock Adversarial examples are not bugs, they are features.
\newblock {\em arXiv preprint arXiv:1905.02175}, 2019.

\bibitem[JBKS15]{jain2015comparative}
Aditya Jain, Ramta Bansal, Avnish Kumar, and KD~Singh.
\newblock A comparative study of visual and auditory reaction times on the
  basis of gender and physical activity levels of medical first year students.
\newblock {\em International Journal of Applied and Basic Medical Research},
  5(2):124, 2015.

\bibitem[KB14]{kingma2014adam}
Diederik~P. Kingma and Jimmy Ba.
\newblock Adam: A method for stochastic optimization, 2014.

\bibitem[KN71]{kubota1971prefrontal}
Kisou Kubota and Hiroaki Niki.
\newblock Prefrontal cortical unit activity and delayed alternation performance
  in monkeys.
\newblock {\em Journal of neurophysiology}, 34(3):337--347, 1971.

\bibitem[KNL06]{kayser2006fixations}
Christoph Kayser, Kristina~J Nielsen, and Nikos~K Logothetis.
\newblock Fixations in natural scenes: Interaction of image structure and image
  content.
\newblock {\em Vision research}, 46(16):2535--2545, 2006.

\bibitem[LBR{\etalchar{+}}15]{luo2015foveation}
Yan Luo, Xavier Boix, Gemma Roig, Tomaso Poggio, and Qi~Zhao.
\newblock Foveation-based mechanisms alleviate adversarial examples.
\newblock {\em arXiv preprint arXiv:1511.06292}, 2015.

\bibitem[LLUZ16]{luo2016understanding}
Wenjie Luo, Yujia Li, Raquel Urtasun, and Richard Zemel.
\newblock Understanding the effective receptive field in deep convolutional
  neural networks.
\newblock In {\em Advances in neural information processing systems}, pages
  4898--4906, 2016.

\bibitem[LMBC90]{land1990eye}
MF~Land, JN~Marshall, D~Brownless, and TW~Cronin.
\newblock The eye-movements of the mantis shrimp odontodactylus scyllarus
  (crustacea: Stomatopoda).
\newblock {\em Journal of Comparative Physiology A}, 167(2):155--166, 1990.

\bibitem[MHG{\etalchar{+}}14]{mnih2014recurrent}
Volodymyr Mnih, Nicolas Heess, Alex Graves, et~al.
\newblock Recurrent models of visual attention.
\newblock In {\em Advances in neural information processing systems}, pages
  2204--2212, 2014.

\bibitem[MLC14]{marshall2014shrimps}
NJ~Marshall, MF~Land, and TW~Cronin.
\newblock Shrimps that pay attention: saccadic eye movements in stomatopod
  crustaceans.
\newblock {\em Philosophical Transactions of the Royal Society B: Biological
  Sciences}, 369(1636):20130042, 2014.

\bibitem[MMS{\etalchar{+}}17]{madry2017towards}
Aleksander Madry, Aleksandar Makelov, Ludwig Schmidt, Dimitris Tsipras, and
  Adrian Vladu.
\newblock Towards deep learning models resistant to adversarial attacks.
\newblock {\em arXiv preprint arXiv:1706.06083}, 2017.

\bibitem[MS85]{messner1985image}
Richard~A Messner and Harold~H Szu.
\newblock An image processing architecture for real time generation of scale
  and rotation invariant patterns.
\newblock {\em Computer vision, graphics, and image processing}, 31(1):50--66,
  1985.

\bibitem[MST85]{massone1985form}
Lina Massone, Giulio Sandini, and Vincenzo Tagliasco.
\newblock “form-invariant” topological mapping strategy for 2d shape
  recognition.
\newblock {\em Computer Vision, Graphics, and Image Processing},
  30(2):169--188, 1985.

\bibitem[MUM83]{mishkin1983object}
Mortimer Mishkin, Leslie~G Ungerleider, and Kathleen~A Macko.
\newblock Object vision and spatial vision: two cortical pathways.
\newblock {\em Trends in neurosciences}, 6:414--417, 1983.

\bibitem[RDS{\etalchar{+}}15]{ILSVRC15}
Olga Russakovsky, Jia Deng, Hao Su, Jonathan Krause, Sanjeev Satheesh, Sean Ma,
  Zhiheng Huang, Andrej Karpathy, Aditya Khosla, Michael Bernstein,
  Alexander~C. Berg, and Li~Fei-Fei.
\newblock {ImageNet Large Scale Visual Recognition Challenge}.
\newblock {\em International Journal of Computer Vision (IJCV)},
  115(3):211--252, 2015.

\bibitem[Sch77]{schwartz1977spatial}
Eric~L Schwartz.
\newblock Spatial mapping in the primate sensory projection: analytic structure
  and relevance to perception.
\newblock {\em Biological cybernetics}, 25(4):181--194, 1977.

\bibitem[She10]{sherrington1910flexion}
Charles~Scott Sherrington.
\newblock Flexion-reflex of the limb, crossed extension-reflex, and reflex
  stepping and standing.
\newblock {\em The Journal of physiology}, 40(1-2):28--121, 1910.

\bibitem[STG12]{Schutz7547}
Alexander~C. Schütz, Julia Trommershäuser, and Karl~R. Gegenfurtner.
\newblock Dynamic integration of information about salience and value for
  saccadic eye movements.
\newblock {\em Proceedings of the National Academy of Sciences},
  109(19):7547--7552, 2012.

\bibitem[SZS{\etalchar{+}}13]{szegedy2013intriguing}
Christian Szegedy, Wojciech Zaremba, Ilya Sutskever, Joan Bruna, Dumitru Erhan,
  Ian Goodfellow, and Rob Fergus.
\newblock Intriguing properties of neural networks.
\newblock {\em arXiv preprint arXiv:1312.6199}, 2013.

\bibitem[TB10]{traver2010review}
V~Javier Traver and Alexandre Bernardino.
\newblock A review of log-polar imaging for visual perception in robotics.
\newblock {\em Robotics and Autonomous Systems}, 58(4):378--398, 2010.

\bibitem[TSSDV82]{tootell1982deoxyglucose}
Roger~B Tootell, Martin~S Silverman, Eugene Switkes, and Russell~L De~Valois.
\newblock Deoxyglucose analysis of retinotopic organization in primate striate
  cortex.
\newblock {\em Science}, 218(4575):902--904, 1982.

\bibitem[UOvdOK18]{uesato2018adversarial}
Jonathan Uesato, Brendan O'Donoghue, Aaron van~den Oord, and Pushmeet Kohli.
\newblock Adversarial risk and the dangers of evaluating against weak attacks,
  2018.

\bibitem[WGB11]{wang2011gateways}
Quanxin Wang, Enquan Gao, and Andreas Burkhalter.
\newblock Gateways of ventral and dorsal streams in mouse visual cortex.
\newblock {\em Journal of Neuroscience}, 31(5):1905--1918, 2011.

\bibitem[WGTB17]{worrall2017harmonic}
Daniel~E Worrall, Stephan~J Garbin, Daniyar Turmukhambetov, and Gabriel~J
  Brostow.
\newblock Harmonic networks: Deep translation and rotation equivariance.
\newblock In {\em Proceedings of the IEEE Conference on Computer Vision and
  Pattern Recognition}, pages 5028--5037, 2017.

\bibitem[WH92]{wilson1992pattern}
JC~Wilson and RM~Hodgson.
\newblock A pattern recognition system based on models of aspects of the human
  visual system.
\newblock In {\em 1992 International Conference on Image Processing and its
  Applications}, pages 258--261. IET, 1992.

\bibitem[XZL{\etalchar{+}}18]{xiao2018spatially}
Chaowei Xiao, Jun-Yan Zhu, Bo~Li, Warren He, Mingyan Liu, and Dawn Song.
\newblock Spatially transformed adversarial examples.
\newblock {\em arXiv preprint arXiv:1801.02612}, 2018.

\bibitem[ZKW99]{zetzsche1999atoms}
Christoph Zetzsche, Gerhard Krieger, and Bernhard Wegmann.
\newblock The atoms of vision: Cartesian or polar?
\newblock {\em JOSA A}, 16(7):1554--1565, 1999.

\end{thebibliography}
\newpage
\appendix
\section{Model Details}
\label{sec:appendix}
\begin{figure}
\includegraphics[width=\textwidth]{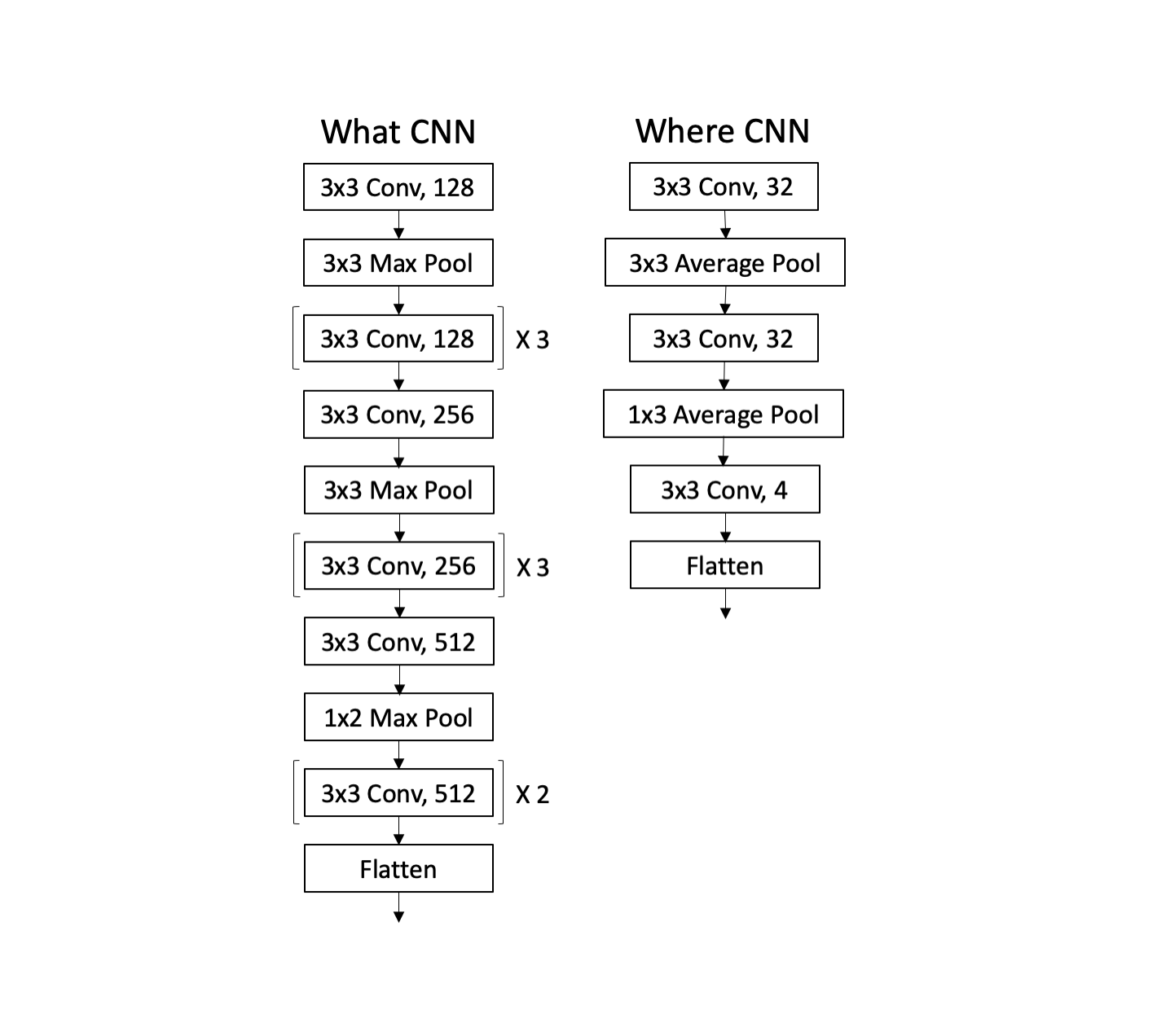}
\caption{\textbf{What and Where pathways for experiment \ref{Training on ImageNet}.} Relu activation, and batch normalization is used after each convolution. Wrap-around padding is used before convolution and pooling layers. These layers are omitted from the figure for clarity.}
\label{fig:imnetarch}
\end{figure}

\begin{figure}
\includegraphics[width=\textwidth]{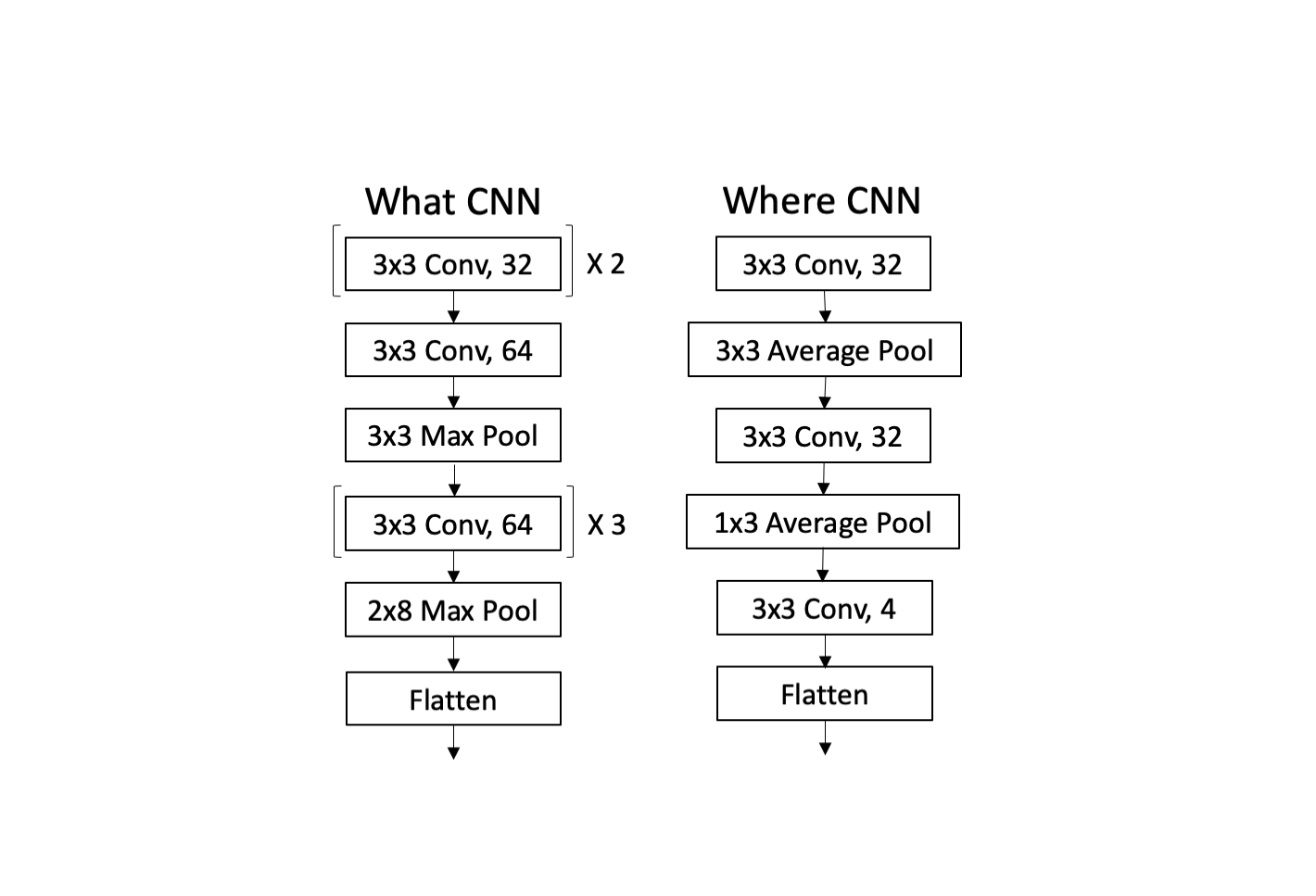}
\caption{\textbf{What and Where pathways for experiment \ref{Rotation and Scale Invariance}.} Relu activation, and batch normalization is used after each convolution. Wrap-around padding is used before convolution and pooling layers. These layers are omitted from the figure for clarity.}
\label{fig:sim2architure}
\end{figure}

The hyperparameters for the experiments \ref{Training on ImageNet}, and \ref{Rotation and Scale Invariance} are summarized in Table \ref{tab:hyperparams}. The architecture of CNNs in "what" and "where" pathways are summarized in Figure \ref{fig:imnetarch}, and \ref{fig:sim2architure}.

\begin{table}[ht]
\caption{\textbf{Hyperparameters.}}
\centering
\begin{tabular}{l c c}
 & section \ref{Training on ImageNet} & section \ref{Rotation and Scale Invariance} \\ 
\hline
$H$ & 54 & 20 \\
$W$ & 108 & 24 \\
$\sigma$ & 0.16 & 0.16 \\
$\rho_{min}$ & $\ln(0.02)$ & $\ln(0.05)$ \\
$\rho_{max}$ & $\ln(1)$ & $\ln(0.8)$ \\
LSTM dim (what path) & 512 & 128\\
LSTM dim (where path) & 512 & 128\\
Learning rate for what path & 1e-4 & 1e-3 \\
Learning rate for where path & 1e-6 & 1e-5 \\
NumGlimpse & 10 & 20\\ 
Batch size & 64 & 256\\
Unit in FC layer (classification net.)&512&128\\
Unit in FC layer (baseline net.)&512&128\\
Unit in FC layer (location net.)&128&128\\
\hline
\end{tabular}
\label{tab:hyperparams}
\end{table}

\end{document}